\PassOptionsToPackage{numbers, compress}{natbib}
\documentclass[sigconf]{acmart}
\renewcommand\footnotetextcopyrightpermission[1]{}
\AtBeginDocument{%
  }
\setcopyright{acmlicensed}
\copyrightyear{2018}
\acmYear{2018}
\acmDOI{XXXXXXX.XXXXXXX}
\acmConference[Preprint]{}{}{}
\usepackage[utf8]{inputenc} %
\usepackage[T1]{fontenc}    %
\usepackage{hyperref}       %
\usepackage{url}            %
\usepackage{booktabs}       %
\usepackage{nicefrac}       %
\usepackage{microtype}      %
\usepackage{enumitem} %
\usepackage{amsmath}
\usepackage{algorithm}
\usepackage{algpseudocode}
\usepackage{multirow}
\usepackage{graphicx}

\usepackage{caption}
\usepackage{subcaption}
\usepackage{svg}
\svgsetup{inkscape=yes}  %

\usepackage[theorems]{tcolorbox} %
\usepackage{makecell}
\tcbuselibrary{listings,theorems,breakable}

\usepackage{fontawesome5}
\usepackage[dvipsnames,table]{xcolor} 
\usepackage{colortbl} 
\usepackage{stfloats}

\newtcbtheorem[number within=subsection]{exmp}{Prompts}%
{breakable,colback=white!5!white,colframe=black!95!,fonttitle=\bfseries, left=.02in, right=.02in,bottom=.02in, top=.02in
,fontupper=\small}{exmp}

\newtcbtheorem[number within=section]{case}{Examples}%
{breakable,colback=white!5!white,colframe=black!95!,fonttitle=\bfseries\small, left=.02in, right=.02in,bottom=.02in, top=.02in ,fontupper=\small}{case}

\newtcbtheorem[number within=section]{case_data}{Examples}%
{breakable,
 colback=white!5!white,
 colframe={rgb,255:red,78; green,136; blue,114}, %
 fonttitle=\bfseries,
 left=.02in,
 right=.02in,
 bottom=.02in,
 top=.02in}{case_data}
\usepackage{xspace}
\newcommand{\model}{NPG-Muse\xspace}
\newcommand{\paratitle}[1]{\vspace{0.95ex}\noindent \textbf{#1}}
\usepackage{pifont}

\begin{document}

\title{\model: Scaling Long Chain-of-Thought Reasoning with NP-Hard Graph Problems}

\author{Yuyao Wang}
\authornotemark[1]
\affiliation{
  \institution{HKUST (GZ)}
  \country{Guangzhou, China}
  }
\email{ywang737@connect.hkust-gz.edu.cn}

\author{Bowen Liu}
\authornote{Equal Contribution.}
\affiliation{
  \institution{HKUST (GZ)}
  \country{Guangzhou, China}
  }
\email{bliu690@connect.hkust-gz.edu.cn}

\author{Jianheng Tang}
\affiliation{
  \institution{HKUST (GZ)}
  \country{Guangzhou, China}
  }
\email{sqrt3tjh@gmail.com}

\author{Nuo Chen}
\affiliation{
  \institution{HKUST (GZ)}
  \country{Guangzhou, China}
  }
\email{chennuo26@gmail.com}

\author{Yuhan Li}
\affiliation{%
  \institution{HKUST (GZ)}
  \country{Guangzhou, China}
  }
\email{yuhanli98@gmail.com}

\author{Qifan Zhang}
\affiliation{%
  \institution{HKUST (GZ)}
  \country{Guangzhou, China}
  }
\email{qzhang297@connect.hkust-gz.edu.cn}

\author{Chenyi Zi}
\affiliation{%
  \institution{HKUST (GZ)}
  \country{Guangzhou, China}
  }
\email{czi447@connect.hkust-gz.edu.cn}

\author{Chen Zhang}
\affiliation{%
  \institution{Createlink Technology}
  \country{Hangzhou, China}
  }
\email{zhangchen@chuanglintech.com}

\author{Jia Li}
\authornote{Corresponding author.}
\affiliation{%
  \institution{HKUST (GZ)}
  \country{Guangzhou, China}
  }
\email{jialee@hkust-gz.edu.cn}

\renewcommand{\shortauthors}{Yuyao Wang et al.}

\begin{abstract}
Reasoning Large Language Models (RLLMs) have recently achieved remarkable progress on complex reasoning tasks, largely enabled by their long chain-of-thought (\textbf{Long CoT}) capabilities. However, developing these Long CoT behaviors relies heavily on post-training with high-quality datasets, which are typically costly and human-curated (e.g., mathematics and code), leaving \textbf{scalable} alternatives unexplored. In this work, we introduce \textbf{NP-hard (NPH) graph problems} as a novel synthetic training corpus, as they inherently require deep reasoning, extensive exploration, and reflective strategies—the core characteristics of Long CoT reasoning. Building on this insight, we develop a two-stage post-training framework: (i) Long-CoT Supervised Fine-Tuning (SFT) on rejection-sampled NPH graph instances, which substantially enhances reasoning depth, and (ii) Reinforcement Learning (RL) with a fine-grained reward design, which sharpens reasoning efficiency. The resulting \model-series models exhibit substantially enhanced Long CoT reasoning capabilities, achieving consistent gains across mathematics, coding, logical, and graph reasoning benchmarks. \model-7B even surpasses QwQ-32B on NPH graph problems in both accuracy and reasoning efficiency. These results position NPH graph problems as an effective and scalable resource for advancing Long CoT reasoning in LLM post-training. Our implementation is available at \url{https://github.com/littlewyy/NPG-Muse}. 
\end{abstract}




\maketitle

\section{Introduction}
Reasoning Large Language Models (RLLMs) \cite{jaech2024openai, openai_o3_2025, openai_o3_mini_2025, openai_o4_mini_2025, guo2025deepseek, comanici2025gemini25pushingfrontier} have demonstrated significant performance leaps in complex reasoning tasks such as mathematics and coding. 
A defining characteristic that distinguishes these RLLMs from traditional LLMs is their long chain-of-thought (\textbf{Long CoT}) capabilities, which enable more detailed and iterative processes of exploration and reflection, thereby enhancing performance on intricate problems \cite{chen2025towards}. As a result, scaling Long CoT capabilities has become a central focus in recent LLM research.

The Long CoT ability in RLLMs is largely attributed to post-training, where Supervised Fine-Tuning (SFT) and Reinforcement Learning (RL) have proven effective \cite{guo2025deepseek}. SFT allows LLMs to imitate advanced reasoning behaviors through step-by-step supervision, while RL encourages self-learning through reward signals, and their combination often yields superior results by balancing guided learning with autonomous exploration. Despite growing interest in post-training strategies, the role of \textbf{training corpora in eliciting Long CoT behavior} remains relatively underexplored. In many challenging domains such as math and coding, high-quality data often rely on costly human curation. Therefore, exploring new training corpora that are both complex and \textbf{synthetically scalable} represents a promising direction.

According to \cite{chen2025towards}, Long CoT behaviors manifest in three critical dimensions: \textbf{(1) Deep Reasoning}, requiring logical depth to sustain extended reasoning chains; \textbf{(2) Extensive Exploration}, involving transitions from known to unknown reasoning paths and generation of uncertain or parallel logic; and \textbf{(3) Feasible Reflection}, encompassing self-correction, refinement, and feedback on logical structures. To induce such behaviors in LLMs, we seek training corpora that either \textit{demonstrate} these patterns (for imitation-based SFT) or \textit{stimulate} them (for self-learning via RL).

We identify \textbf{NP-hard (NPH) graph problems} as particularly well-suited for this purpose. NPH graph problems possess several advantageous properties that align naturally with the core characteristics of Long CoT reasoning—namely, \textbf{Deep Reasoning}, \textbf{Extensive Exploration}, and \textbf{Feasible Reflection}. We examine each dimension in turn: \ding{182} \textit{Deep Reasoning.} Due to the exponential worst-case complexity of NP-hard graph problems, their resolution through natural language simulation in Large Language Models (LLMs) may necessitate extensive reasoning chains even for small-scale instances. This characteristic renders such problems ideally suited for eliciting deep reasoning. \ding{183} \textit{Extensive Exploration.} The absence of known polynomial-time solutions forces LLMs to autonomously explore unknown optimal strategies, fostering extensive exploration. This exploration is further enriched by the intrinsic complexity of graph problems, which elicits diverse reasoning patterns, including topological analysis, logical deduction, enumeration, precise computation, and case decomposition \cite{Zhang2025Improving}. \ding{184} \textit{Feasible Reflection.} Due to the diversity of graph structures and the computational feasibility of exact solvers for small-scale NP-hard problems, these tasks facilitate the scalable synthesis of datasets with verifiable ground truth. Consequently, this enables large-scale post-training and provides a robust basis for feasible reflection. To empirically validate that NPH graph problems can elicit substantially longer reasoning chains, we apply QwQ-32B \cite{qwq32b} to solve small-scale NPH graph problem instances. After rejection sampling based on final answer correctness, the resulting samples exhibit an \textbf{average length exceeding 11k tokens}—more than \textbf{2.3$\times$ longer} than the widely used high-quality Long CoT dataset s1K \cite{Muennighoff2025Simple}, which averages 4.7k tokens. This substantial length difference empirically demonstrates the superior potential of NPH graph problems as a source for Long CoT training data.

Building on these insights, we propose a novel two-stage framework that leverages NPH graph problems to enhance the Long CoT capabilities of LLMs. In the first stage, we perform \textbf{Long-CoT Supervised Fine-Tuning (SFT)} on base models (Qwen2.5-7B-Instruct-1M \cite{qwen2.5-1m} and Qwen3-8B-Base \cite{qwen3}) using \textbf{small-scale NPH graph problems} generated by prompting QwQ-32B \cite{qwq32b} and applying rejection sampling based on final answer correctness. This stage substantially enhances the models’ response length and reasoning performance across diverse complex tasks. However, we also observe that the models exhibit severe reasoning redundancy, which not only undermines reasoning efficiency but also leads to significant performance degradation on certain tasks. This motivates the second training stage, where we apply \textbf{Reinforcement Learning (RL)} to promote more efficient and broadly applicable Long CoT reasoning. We design a fine-grained, outcome-based reward function that incorporates: (1) a repetition penalty to directly reduce redundant reasoning, (2) a graph-specific solution quality reward to encourage concise yet effective solutions, and (3) a format reward to promote structured reasoning. This stage produces our \model-series models: \model-7B and \model-8B.

Our \model-series models demonstrate significantly enhanced Long CoT capabilities, as evidenced by strong generalization across diverse complex reasoning tasks along two dimensions:  
(1) \textit{In-task generalization}: Although trained exclusively on small-scale NP-hard graph problems, both models achieve remarkable accuracy improvements on large-scale instances, indicating that they learn transferable reasoning strategies rather than memorizing specific solutions.  
(2) \textit{Cross-domain generalization}: both models attain notable gains across other complex reasoning domains, including mathematics, coding, logic, and graph reasoning.

Beyond these explicit improvements on benchmark accuracy, we also uncover implicit evidence that \model-series models have genuinely advanced their Long CoT capabilities:  (1) a substantial increase in response length, aligning with the \textit{Deep Reasoning} characteristic of Long CoT;  
(2) significant improvements in \textit{pass@k}, suggesting a more diverse exploration space consistent with \textit{Extensive Exploration};  
(3) a marked rise in reflection frequency, corresponding to the \textit{Feasible Reflection} feature of Long CoT. 

In summary, our main contributions are as follows:
\begin{itemize}[leftmargin=*, topsep=4pt]
\item \textbf{Novel Corpus.} We are the first to introduce NP-hard graph problems as a training corpus for eliciting Long Chain-of-Thought capabilities in LLMs.

\item \textbf{Training Pipeline.} We propose a two-stage post-training framework that combines Long-CoT SFT guided by rejection-sampled NP-hard graph instances, and RL with a fine-grained reward design, fostering effective and efficient Long CoT reasoning.

\item \textbf{Superior Performance.} The resulting \model-series models exhibit substantially enhanced Long CoT reasoning, achieving gains across mathematics, coding, logical, and graph reasoning benchmarks. \model-7B even surpasses QwQ-32B on NP-hard graph problems despite its significantly smaller size.

\end{itemize}

\begin{figure*}[t]
  \centering
  \includegraphics[width=\textwidth]{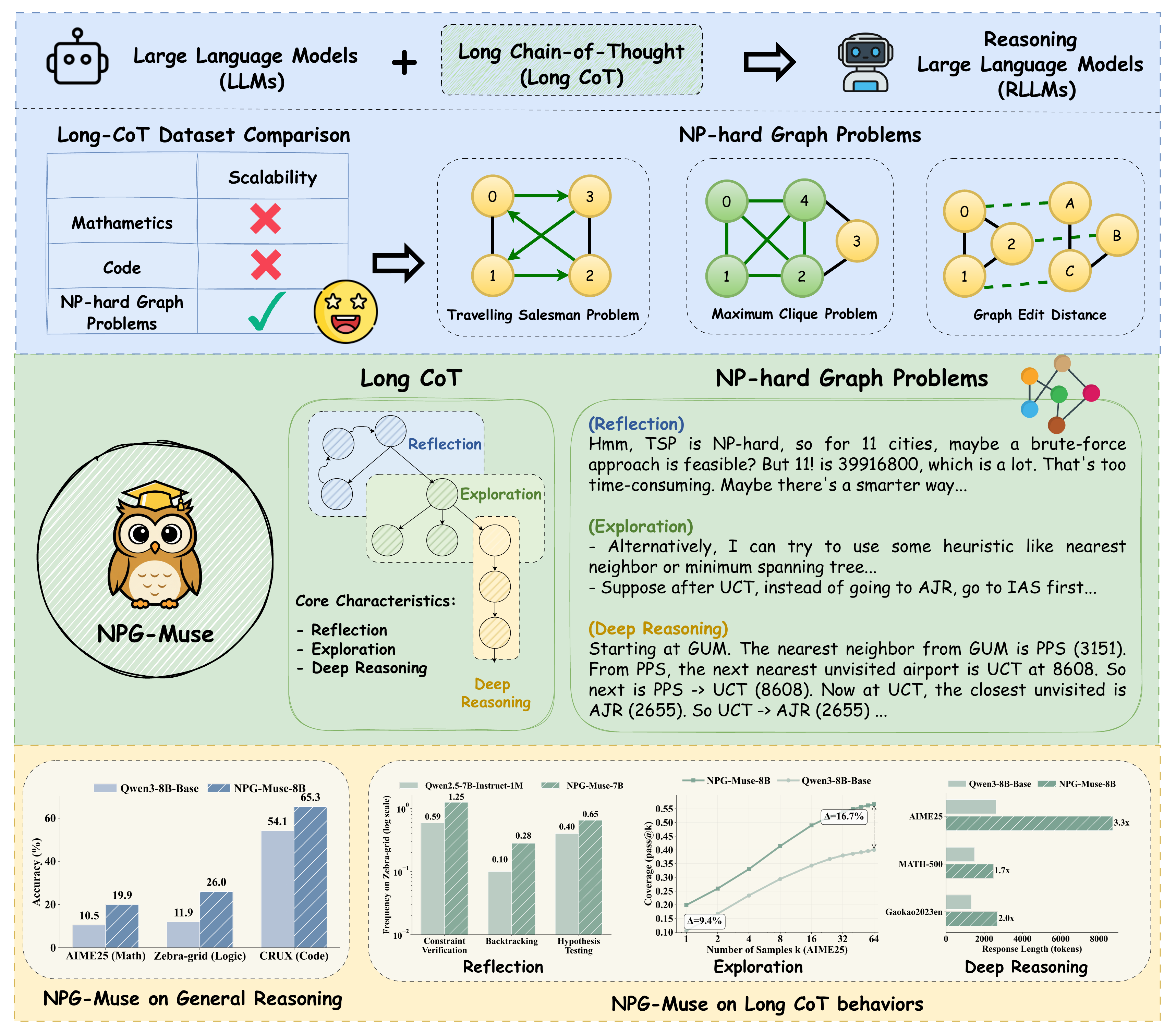}
  \caption{Overview of \model.}
  \label{fig:wide}
\end{figure*}
\section{Preliminaries}
\subsection{NP-Hard Graph Problems}

NP-hard (NPH) graph problems are intractable challenges in theoretical computer science with no known polynomial-time solutions. 
We select three representative NPH graph problems for training: Graph Edit Distance (GED), Traveling Salesman Problem (TSP), and Maximum Clique Problem (MCP). These tasks are chosen for their structural and operational diversity: GED evaluates the similarity between graph pairs via edit operations, TSP identifies optimal paths in weighted graphs, and MCP seeks dense substructures in unweighted graphs. This diversity minimizes overlap in solution strategies, thereby introducing a rich variety of reasoning paradigms. The formal definitions of these tasks are as follows:

\begin{itemize}[leftmargin=*]
\item \textbf{Graph Edit Distance (GED).} Given two graphs \(G_1 = (V_1, E_1)\) and \(G_2 = (V_2, E_2)\), GED is the minimum cost of a sequence of edit operations (insertion, deletion, or substitution of nodes or edges) that transforms \(G_1\) into \(G_2\) \cite{serratosa2021redefining}.
\item \textbf{Traveling Salesman Problem (TSP).} Given a weighted graph \(G = (V, E)\), TSP asks for the shortest possible cycle that visits each vertex exactly once and returns to the starting point \cite{dantzig1954solution}.
\item \textbf{Maximum Clique Problem (MCP).} Given an unweighted graph \(G = (V, E)\), MCP seeks the largest subset of vertices \(C \subseteq V\) such that every pair of vertices in \(C\) is connected by an edge \cite{bomze1999maximum}.
\end{itemize}
\subsection{Graph Serialization Format}

To facilitate natural language reasoning for NP-hard graph problems, we serialize graph structures into a human-readable text format following GraphArena \cite{tang2024grapharena}. A condensed example is provided below, while the full templates are detailed in Appendix \ref{app:implementation}.
\begin{exmp}{A condensed example of TSP for graph serialization format}{exmp:simple}
- Airports to visit: BVC, URA, SMR, KOK, MTV, TFF, FLN, YZF\\
- Travel distances (in kilometers) between each pair of airports:\\
BVC to KOK: 6401\\
FLN to YZF: 11826 
\end{exmp}

\section{Methodology}
Our method consists of two stages: Long-CoT Supervised Fine-Tuning (SFT) and Reinforcement Learning (RL). The SFT stage first equips the model with Long CoT reasoning capabilities, while the RL stage is further applied to enhance reasoning efficiency.

\begin{figure*}[htbp]
  \centering
  \includegraphics[width=\textwidth]{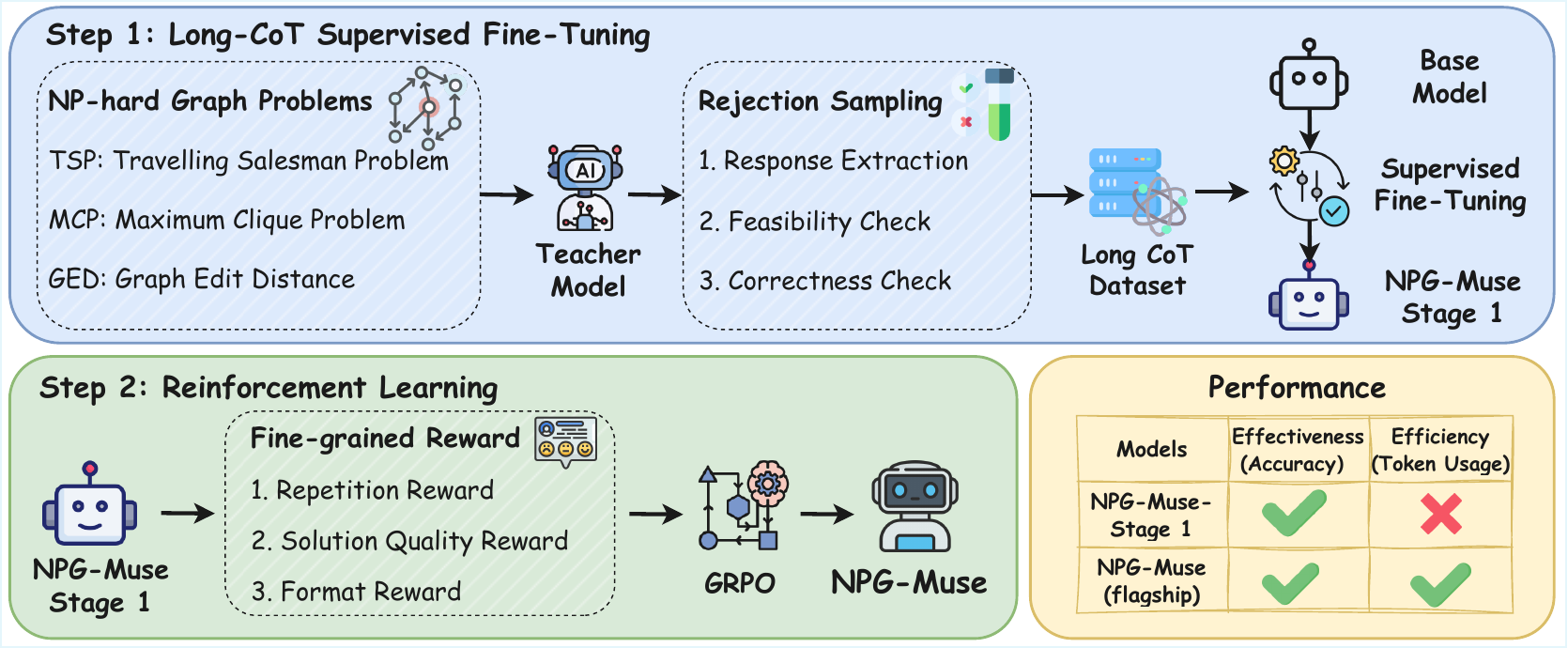}
  \caption{Methodology of \model.}
  \label{fig:method}
\end{figure*}

\subsection{Long-CoT Supervised Fine-Tuning}
Long-CoT Supervised Fine-Tuning (SFT) comprises two components: (1) constructing a Long CoT reasoning dataset via distillation, and (2) training the model through SFT on this dataset to acquire a systematic step-by-step long-reasoning paradigm.

\subsubsection{Long CoT Dataset Construction}
\label{sec:rejection-sampling}
We construct a Long CoT dataset comprising 9,000 samples by prompting QwQ-32B with NP-hard graph problems and applying \textbf{rejection sampling} \cite{rejectionsampling} to retain only those with correct answers. Each sample is a QA tuple \((G, x, r, y)\), where \(G\) denotes the graph structure, \(x\) is the task-specific question, \(r\) contains the CoT reasoning path, and \(y\) is the final answer. The rejection sampling process is adapted from the validation pipeline of GraphArena, ensuring that only well-formed, feasible, and correct solutions are preserved through a multi-round quality control procedure:

\vspace{7pt}
\noindent \textbf{1. Response Extraction}: We first parse \textbf{structured solutions} rather than purely numerical answers from the LLM outputs using regular expressions. Responses with formatting errors that prevent successful parsing are immediately discarded as infeasible. For NP-hard graph problem evaluation, structured solutions provide finer granularity and higher reliability compared to numerical answers. The expected formats are:
\begin{itemize}[leftmargin=*, topsep=2pt]
\item \textbf{TSP}: A list representing the visiting route.
\item \textbf{GED}: A list specifying the node mapping between two graphs.
\item \textbf{MCP}: A list of node IDs in the clique.
\end{itemize}

\vspace{7pt}
\noindent \textbf{2. Feasibility Check}: Parsed solutions are then verified against task-specific constraints to ensure logical validity:
\begin{itemize}[leftmargin=*, topsep=2pt]
\item \textbf{TSP}: The tour must visit each node once and return to start. 
\item \textbf{GED}: The node mapping must cover all nodes without repetition.
\item \textbf{MCP}: The selected nodes must form a complete subgraph (clique).
\end{itemize}

\vspace{7pt}
\noindent \textbf{3. Correctness Check}: Solutions that pass the feasibility check are further validated by an algorithmic verifier. Specifically, instead of directly checking the numerical answers, we validate the detailed solution structures through algorithmic verification:
\begin{itemize}[leftmargin=*, topsep=2pt]
\item \textbf{TSP}: The constructed tour must yield a total path length equal to the ground-truth value.
\item \textbf{GED}: The computed edit distance must be equal to the ground truth.
\item \textbf{MCP}: The identified clique must have the same size as the known maximum.
\end{itemize}

Only responses that pass all three stages are retained, so the SFT dataset comprises exclusively well-structured, feasible, and correct examples. At the same time, although all training instances are drawn from small-scale graphs, problem complexity—measured by the number of nodes—is carefully controlled to ensure a balanced distribution across levels. This construction simultaneously secures the \textbf{quality} and \textbf{difficulty} of the Long CoT dataset, providing a reliable foundation for the subsequent SFT stage.

\subsubsection{Supervised Fine-Tuning}
Following dataset construction, we conduct Supervised Fine-Tuning (SFT) to endow the model with Long CoT reasoning capabilities. 
Given a graph \( G \) and a question \( x \) as input, the model is trained to generate a reasoning path \( r \) and subsequently the final answer \( y \). 
We adopt a standard language modeling objective, defined as:
\begin{equation}
\mathcal{L}_{\text{SFT}} = -\sum_{i=1}^N \sum_{j=1}^{M_i} \log P(r_{i,j}, y_i \mid G_i, x_i; \theta),
\end{equation}
where \( N \) is the total number of QA pairs, \( M_i \) denotes the number of reasoning steps in the CoT sequence of the \( i \)-th sample, and \( \theta \) represents the model parameters. 

This stage not only drives the model to produce long reasoning chains capable of solving NP-hard graph problems, but also implicitly encourages the development of systematic, step-by-step reasoning skills for tackling general complex reasoning tasks.

\subsection{Reinforcement Learning}
The goal of the RL stage is to address the reasoning inefficiency that emerges after the Long CoT SFT stage. By designing fine-grained reward functions, we guide the model to reduce redundant reasoning while maintaining solution quality, thereby producing more efficient reasoning paths. We employ Group Relative Policy Optimization (GRPO) \cite{shao2024deepseekmath} as our RL algorithm, chosen for its strong performance and low memory requirement.

Our reward function targets reasoning efficiency from three complementary perspectives.  
First, we directly penalize specific redundancy patterns in reasoning, which we implement via the \textit{repetition reward}.  
Second, we implicitly promote efficient reasoning by rewarding high-quality solutions generated within limited context window, implemented as the \textit{solution quality reward}.  
Third, we encourage outputs that follow structured format to reduce disorganized reasoning, implemented via the \textit{format reward}.  
Formally, the reward function consists of three core components: \textit{repetition reward}, \textit{solution quality reward}, and \textit{format reward}.

\vspace{10pt}
\paratitle{Repetition Reward. } Addressing overthinking in Long CoT reasoning has attracted substantial research attention. 
However, because overthinking is highly context-dependent, developing efficient and robust algorithms for its effective handling remains challenging. Consequently, many RL-based approaches adopt a simpler proxy, leveraging a response length reward to indirectly discourage overthinking by penalizing excessively long responses \cite{Yue2024Dont}.
While this approach can reduce response length, it may also constrain the model’s ability to perform deeper reasoning, since reasoning length does not necessarily correlate with overthinking.
We therefore focus on reducing overthinking based on the content of the reasoning process rather than its length. We identify a prevalent and detrimental overthinking pattern—\emph{repetition}—and develop an efficient algorithm for its detection.  
Repetition refers to the phenomenon that LLM repeatedly generates the same statements until it reaches the context length limit.  
Such repetition not only undermines reasoning efficiency but also disrupts logical progression, often leading to low-quality solutions. 
Specifically, we define repetition as any substring of at least 20 characters repeated consecutively 5 times within a model response.  
This parameter setting, derived from empirical observations, effectively captures genuine repetition while avoiding false positives on valid reasoning.  
Specifically, we adopt a linear-time repetition detection algorithm based on suffix trees \cite{blumer1985, CROCHEMORE1981244} (see Appendix~\ref{app:repetition}). 
When repetition is detected, we apply a fixed penalty of -1 to the total reward.

\vspace{10pt}
\paratitle{Solution Quality Reward.} We evaluate the quality of solutions within the context window, thereby encouraging reasoning that is concise yet effective. 
This evaluation adheres to the principles established in the Long CoT dataset construction (see in Section \ref{sec:rejection-sampling}. Specifically, model responses are categorized into three types: \emph{optimal}, \emph{suboptimal}, and \emph{hallucinated}.  
Optimal solutions yield the exact correct answer and receive the highest reward of 2.0.  Hallucinated solutions that violate fundamental problem constraints are penalized with a reward of -1 to discourage unproductive exploration by the model. Suboptimal solutions satisfy the core problem constraints but fail to achieve optimality, often due to premature termination or inefficient exploration. For these, we assign a reward scaled between 0 and 0.5 based on the ratio between the achieved objective value and the optimal value. Specifically, the suboptimal reward is assigned as follows (see Appendix~\ref{app:solution} for more details).

\begin{itemize}[leftmargin=*, topsep=2pt]
    \item \textbf{TSP}: $R_{\textnormal{sub}} = \left(\tfrac{ans}{d}\right)^2 \times 0.5$, where $d$ denotes the predicted tour length.
    \item \textbf{GED}: $R_{\textnormal{sub}} = \tfrac{ans}{c} \times 0.5$, where $c$ denotes the predicted edit cost.
    \item \textbf{MCP}: $R_{\textnormal{sub}} = \tfrac{s}{ans} \times 0.5$, where $s$ denotes the predicted clique size.
\end{itemize}

\vspace{10pt}

\paratitle{Format Reward. } We assign a format reward of 1.0 if the model response begins with the token "<think>".  This encourages the model to show explicit reasoning phase before producing the final answer, thereby reducing disorganized thought patterns.
\section{Experiments}
\subsection{Experimental Protocols}
In this section, we conduct extensive experiments to evaluate \model, guided by the following questions: 
-\textbf{RQ1:} How does \model perform on NP-hard graph problems?
-\textbf{RQ2:} How does \model generalize to unseen complex reasoning tasks? 
-\textbf{RQ3:} Does \model truly demonstrate the Long CoT capabilities?

\vspace{-5pt}
\subsubsection{Datasets and Metrics}\label{sec:dataset}
The evaluated datasets fall into two categories: 
(1) \textbf{Seen NP-hard graph problems} from the training stage, including TSP, GED, and MCP provided by GraphArena \cite{tang2024grapharena}. 
The evaluation covers two settings: \emph{small-scale} and \emph{large-scale} problems. Notably, only small-scale samples are used for training, while large-scale datasets evaluate the model’s transferability and practicality.
For GED and TSP, the small and large settings involve graphs with $4$--$9$ and $10$--$20$ nodes; for MCP, the corresponding scales are $4$--$14$ and $15$--$30$ nodes. (2) \textbf{Unseen complex reasoning problems} for testing the model's generalization ability, including AIME24/25 \cite{aime}, MATH-500 \cite{math500}, GaoKao2023en \cite{gaokao2023}, GSM8K \cite{cobbe2021gsm8k},  College Math \cite{collegemath}, ASDIV \cite{miao2020asdiv}, SVAMP \cite{patel2021svamp} and TabMWP \cite{lu2022tabmwp} for mathematical reasoning; CRUX \cite{gu2024cruxevalbenchmarkcodereasoning} for code reasoning; Zebra-grid \cite{lin2025zebralogicscalinglimitsllms} for logical reasoning; GraphInstruct \cite{chen2024graphwiz} and problems unseen during training from GraphArena~\cite{tang2024grapharena} for graph reasoning.  

For the seen NP-hard graph problems, the evaluation metrics include \emph{accuracy} and \emph{feasibility}. Accuracy measures the probability of obtaining the optimal solution, and feasibility measures the probability that the solution satisfies the basic problem constraints. 
For most unseen complex reasoning problems, we adopt the evaluation metrics used in their original benchmarks, which are primarily single-test accuracy. For the exceptionally challenging mathematical datasets AIME24/25, MATH-500 and GaoKao2023en, we employ \textit{avg@k} and \textit{pass@k} to provide a more robust assessment.  The \textit{avg@k} metric calculates the mean accuracy over $k$ independent attempts, reflecting the model's average-case stability; conversely, \textit{pass@k} measures the probability of achieving at least one correct solution within $k$ samples, characterizing the model's potential coverage of the solution space. To balance evaluation comprehensiveness with computational efficiency, we set $k=64$ for AIME24/25 and $k=8$ for MATH-500 and GaoKao2023en. Further details on the datasets and metrics are provided in Appendix~\ref{app:datasets}.

\vspace{-5pt}
\subsubsection{Baselines}  
For comparison on NP-hard graph problems, we consider both general-purpose language models and graph-oriented models.  
The general-purpose models include: (i) closed-source models, GPT-4o \cite{GPT4OpenAI} and Claude-3.5-Sonnet \cite{claude35sonnet}; (ii) open-source reasoning models, QwQ-32B \cite{qwq32b} and S1.1-7B \cite{muennighoff2025s1}
; and (iii) open-source non-reasoning models Llama3-70B-Instruct \cite{llama3modelcard} and our base models (Qwen3-8B-Base \cite{qwen3} and Qwen2.5-7B-Instruct-1M \cite{qwen2.5-1m}).  
The graph-oriented models include Graphwiz-7B-DPO \cite{chen2024graphwiz}, Llama-8B-GT \cite{graphthought} and G1-7B \cite{g1}. For evaluation on unseen reasoning tasks, we adopt the base models of our \model-8B and \model-7B, namely Qwen3-8B-Base and Qwen2.5-7B-Instruct-1M. We additionally include larger models from the same family, Qwen3-14B-Base and Qwen2.5-14B-Instruct-1M, as scaling baselines. Additional details about these baselines are provided in Appendix~\ref{app:baselines}.

\vspace{-5pt}
\subsubsection{Implementation Details}\label{sec:implementation}

We fine-tune Qwen3-8B and Qwen2.5-7B-Instruct-1M to obtain \model-8B and \model-7B, respectively, using PyTorch on 8 NVIDIA A800 (80GB) GPUs. Training is conducted with the Verl framework~\cite{sheng2024hybridflow}.
In the Long-CoT SFT stage, we train for two epochs with a batch size of 8, a cosine learning rate scheduler with an initial learning rate of $1\times10^{-5}$ and a warmup ratio of 0.1, and a context length of 128k tokens.
In the RL stage, we adopt the GRPO algorithm with a batch size of 128, a PPO mini-batch size of 32, a rollout number of 8, a sequence parallel size of 4, a KL coefficient of $1\times10^{-3}$, a learning rate of $1\times10^{-6}$, a maximum response length of 8192, and a temperature of 1.0. More implementation details are provided in Appendix~\ref{app:implementation}.
\subsection{Main Results}
\subsubsection{Effect on NP-hard Graph Problems (RQ1)} 
We evaluate the \model-series on NP-Hard graph problems, benchmarking against base models as well as state-of-the-art reasoning- and graph-oriented language models under two settings: (1) small-scale graphs consistent with training, and (2) large-scale graphs that test generalization and real-world applicability.

\begin{figure}[htbp]
\centerline{\includegraphics[width=0.95\columnwidth]{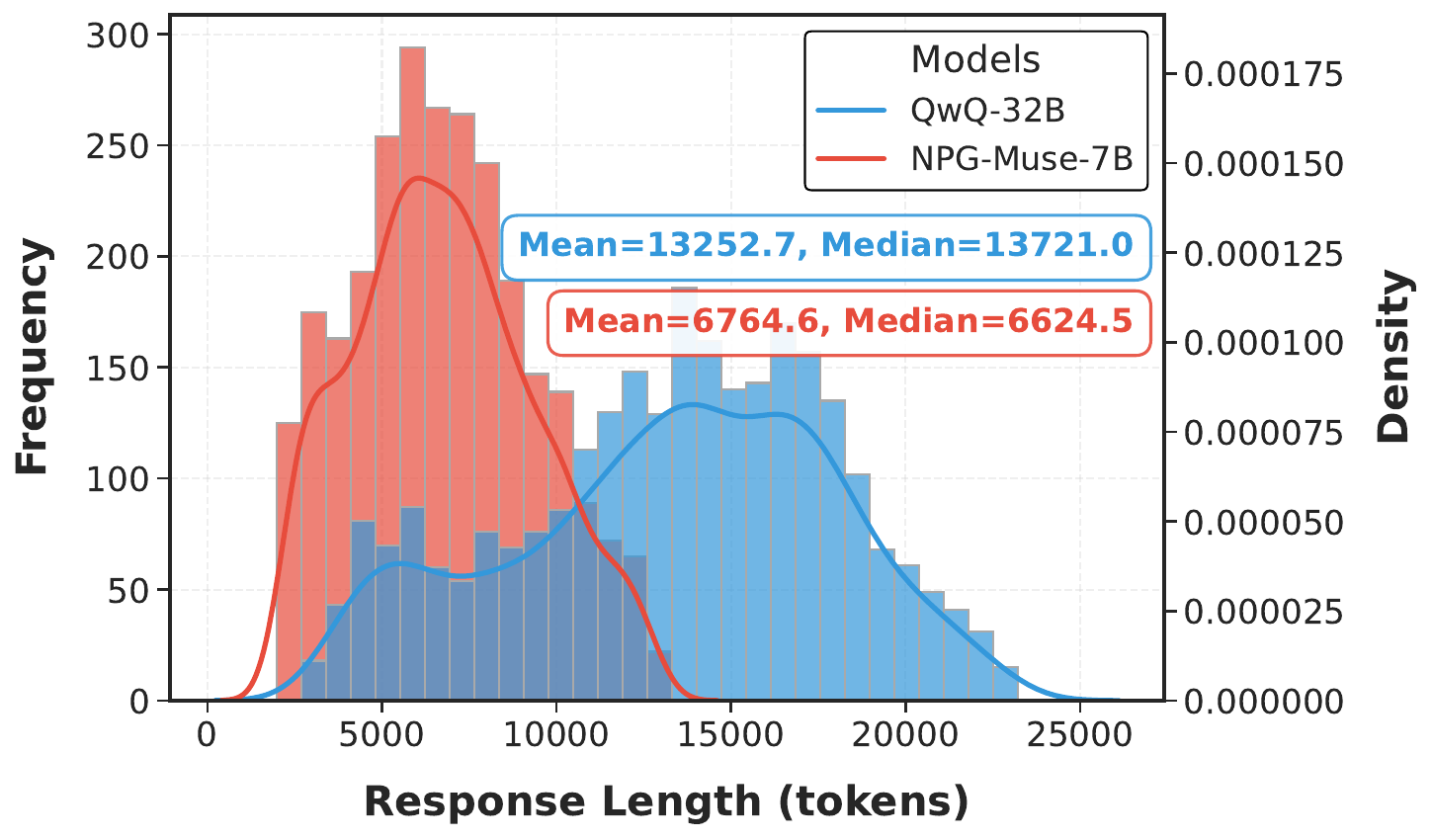}}
    \caption{Token efficiency compared to QwQ-32B.}
    \label{fig:token_main}
\end{figure}

Results are reported in Table~\ref{tab:baseline-comparison-small}.
\model-7B achieves the superior overall accuracies of 86.7\% and 28.1\% on small- and large-scale graphs, respectively. Meanwhile, \model-8B ranks third, performing marginally below QwQ-32B.
Notably, although trained only on small-scale tasks, our models generalize effectively to large-scale ones. For example, \model-7B attains 28.1\% accuracy on large graphs—a 17$\times$ improvement over its base model (1.6\%)—and even outperforms its teacher model QwQ-32B. These results confirm genuine generalization rather than memorization, validating the effectiveness of our approach in stimulating complex reasoning. Compared with closed-source baselines, which yield slightly higher feasibility on large graphs, our models achieve markedly higher accuracy. This reflects a trade-off: closed-source models appear to emphasize constraint propagation to secure feasible outputs, whereas ours prioritize deeper exploration of the solution space for accuracy gains. Beyond accuracy, the \model series also exhibits strong token efficiency. 
As shown in Figure~\ref{fig:token_main}, \model-7B requires only 6764 tokens on average, a 49\% reduction compared to QwQ-32B (13252 tokens), highlighting the favorable balance between accuracy and efficiency achieved by our models.

\begin{table}[htbp]
\caption{Performance comparison on small-scale TSP, GED, and MCP tasks. "Acc." and "Fea." denote Accuracy and Feasibility (both in \%). Best results are \textbf{bold}; second-best are \underline{underlined}. $^\dagger$Qwen2.5-7B-Ins-1M refers to Qwen2.5-7B-Instruct-1M.}
\label{tab:baseline-comparison-small}
\centering
\scriptsize
\setlength{\tabcolsep}{2.5pt}
\begin{tabular}{l|l|cc|cc|cc|cc}
\toprule
\multirow{2}{*}[-0.5ex]{\textbf{Category}} & \multirow{2}{*}[-0.5ex]{\textbf{Model}} & \multicolumn{2}{c|}{\textbf{TSP}} & \multicolumn{2}{c|}{\textbf{GED}} & \multicolumn{2}{c|}{\textbf{MCP}} & \multicolumn{2}{c}{\textbf{Average}} \\
\cmidrule(lr){3-4} \cmidrule(lr){5-6} \cmidrule(lr){7-8} \cmidrule(lr){9-10}
& & \textbf{Acc.} & \textbf{Fea.} & \textbf{Acc.} & \textbf{Fea.} & \textbf{Acc.} & \textbf{Fea.} & \textbf{Acc.} & \textbf{Fea.} \\
\midrule
\multirow{2}{*}{Closed-source} & Claude-3.5-sonnet & 45.4 & \textbf{100.0} & 37.2 & 94.2 & 62.2 & 88.0 & 48.3 & 94.1 \\
& GPT-4o & 44.2 & \underline{99.8} & 32.6 & \underline{98.6} & 62.4 & 79.6 & 46.4 & 92.7 \\
\midrule
\multirow{2}{*}{Reasoning} & QwQ-32B & \underline{89.4} & \textbf{100.0} & \underline{70.2} & \textbf{99.2} & \textbf{96.2} & \underline{97.0} & \underline{85.3} & \textbf{98.7} \\
& S1.1-7B & 33.8 & 56.4 & 22.8 & 81.0 & 46.2 & 80.4 & 34.3 & 72.6 \\
\midrule
\multirow{3}{*}{Non-reasoning} 
& LLaMA3-70B-Ins & 23.2 & \underline{99.8} & 31.6 & 97.0 & 42.8 & 44.2 & 32.5 & 80.3 \\
& Qwen3-8B-Base & 24.8 & 68.4 & 23.4 & 62.0 & 22.4 & 52.8 & 23.5 & 61.1 \\
& Qwen2.5-7B-Ins-1M$^\dagger$ & 19.2 & 62.6 & 23.8 & 72.8 & 34.0 & 60.2 & 25.7 & 65.2 \\
\midrule
\multirow{3}{*}{Graph-oriented} 
& G1-7B & 38.0 & / & 52.0 & / & 30.0 & / & 40.0 & / \\
& LLaMA-8B-GT & 39.2 & / & 46.0 & / & 95.2 & / & 60.1 & / \\
& GraphWiz-7B-DPO & 0.4 & 1.0 & 2.4 & 17.0 & 1.6 & 39.6 & 1.5 & 19.2 \\
\midrule
\multirow{2}{*}{\model} 
& \model-8B \textit{(Ours)} & 82.2 & 99.0 & 63.0 & \textbf{99.2} & 93.6 & \textbf{97.8} & 79.6 & \textbf{98.7} \\
& \model-7B \textit{(Ours)} & \textbf{91.2} & 98.6 & \textbf{72.8} & 98.0 & \underline{96.0} & \textbf{97.8} & \textbf{86.7} & \underline{98.1} \\
\bottomrule
\end{tabular}
\end{table}

\begin{table}[htbp]
\caption{Performance comparison on large-scale TSP, GED, and MCP tasks.}
\label{tab:baseline-comparison-large}
\centering
\scriptsize
\setlength{\tabcolsep}{2.5pt}
\begin{tabular}{l|l|cc|cc|cc|cc}
\toprule
\multirow{2}{*}[-0.5ex]{\textbf{Category}} & \multirow{2}{*}[-0.5ex]{\textbf{Model}} & \multicolumn{2}{c|}{\textbf{TSP}} & \multicolumn{2}{c|}{\textbf{GED}} & \multicolumn{2}{c|}{\textbf{MCP}} & \multicolumn{2}{c}{\textbf{Average}} \\
\cmidrule(lr){3-4} \cmidrule(lr){5-6} \cmidrule(lr){7-8} \cmidrule(lr){9-10}
& & \textbf{Acc.} & \textbf{Fea.} & \textbf{Acc.} & \textbf{Fea.} & \textbf{Acc.} & \textbf{Fea.} & \textbf{Acc.} & \textbf{Fea.} \\
\midrule
\multirow{2}{*}{Closed-source} 
& Claude-3.5-sonnet & 2.8 & \underline{99.6} & 3.4 & 83.6 & 19.0 & 79.8 & 8.4 & \underline{87.7} \\
& GPT-4o & 0.8 & 95.6 & 2.4 & \textbf{97.2} & 15.2 & 63.2 & 6.1 & 85.3 \\
\midrule
\multirow{2}{*}{Reasoning} 
& QwQ-32B & \underline{7.6} & \textbf{100.0} & \textbf{7.4} & 94.0 & \underline{64.6} & 74.6 & \underline{26.5} & \textbf{89.5} \\
& S1.1-7B & 0.0 & 21.0 & 1.0 & 42.0 & 2.2 & 51.0 & 1.1 & 38.0 \\
\midrule
\multirow{3}{*}{Non-reasoning} 
& LLaMA3-70B-Ins & 0.0 & 84.6 & 1.8 & 82.0 & 10.6 & 16.4 & 4.1 & 61.0 \\
& Qwen3-8B-Base & 0.0 & 33.2 & 2.0 & 45.0 & 1.6 & 35.4 & 1.2 & 37.9 \\
& Qwen2.5-7B-Ins-1M$^\dagger$ & 1.2 & 58.8 & 1.2 & 58.8 & 2.4 & 30.2 & 1.6 & 49.3 \\
\midrule
\multirow{3}{*}{Graph-oriented} 
& G1-7B & 0.0 & / & 3.0 & / & 3.0 & / & 2.0 & / \\
& LLaMA-8B-GT & 3.6 & / & 0.8 & / & 63.4 & / & 22.6 & / \\
& GraphWiz-7B-DPO & 0.0 & 0.0 & 0.0 & 18.0 & 0.0 & 50.8 & 0.0 & 22.9 \\
\midrule
\multirow{2}{*}{\model} 
& \model-8B \textit{(Ours)} & 4.8 & 75.2 & \underline{6.0} & \underline{97.0} & 57.2 & \textbf{83.0} & 22.7 & 85.1 \\
& \model-7B \textit{(Ours)} & \textbf{9.4} & 62.0 & 5.6 & 91.0 & \textbf{69.2} & \underline{82.2} & \textbf{28.1} & 78.4 \\
\bottomrule
\end{tabular}
\end{table}

\subsubsection{Effect on Unseen Complex Reasoning Tasks (RQ2)}

Table~\ref{tab:all_performance} reports the generalization performance of our models on mathematical, logical, code, and unseen graph reasoning benchmarks, demonstrating a consistent plug-and-play benefit of the proposed post-training pipeline.

Quantitatively, gains are observed for both the 7B and 8B families. On general reasoning benchmarks, \model-7B increases macro accuracy from 65.6\% to 68.0\%, while \model-8B rises from 63.2\% to 74.1\%. For challenging mathematical tasks evaluated with \textit{avg@k}, the 7B backbone shows steady improvements (34.9\% to 36.4\%), whereas the 8B backbone exhibits a large jump (35.9\% to 50.9\%), with \model-8B outperforming larger 14B scaling baselines. On unseen graph reasoning tasks, despite training only on small-scale TSP, GED, and MCP instances, transfer remains strong: \model-7B improves the graph reasoning average from 32.9\% to 36.2\%, and \model-8B increases from 25.2\% to 54.4\%, with gains across linear, polynomial-time, and NP-class problems. These results indicate that the method strengthens genuine symbolic and structural reasoning rather than relying on memorization.

Importantly, the direction of improvement is consistent across backbones, although the magnitude varies. Smaller backbones obtain modest gains, while stronger backbones yield substantially larger uplift from the same procedure. Together, these findings demonstrate the effectiveness and broad applicability of NP-hard graph problems as a scalable training resource for enhancing complex reasoning capabilities.

\begin{table}[htbp]
\centering
\caption{Performance comparison on reasoning benchmarks. 
The best results are shown in \textbf{bold}, the second-best results are \underline{underlined}, 
and highlighted cells indicate the better-performing model within each base–ours pair. 
$^\dagger$Qwen2.5-14B-Ins-1M denotes Qwen2.5-14B-Instruct-1M. 
Avg.\ denotes macro-average across tasks.}
\label{tab:all_performance}

\definecolor{custompurple}{RGB}{208,224,239}
\scriptsize
\textbf{(a) General Reasoning}
\vspace{2pt}

\setlength{\tabcolsep}{1.5pt}
\begin{tabular}{lccccc|c|c|c}
\toprule
\multirow{2}{*}{Models} & \multicolumn{5}{c}{Mathematics} & Logic & Code & \multirow{2}{*}{Avg.} \\
\cmidrule(lr){2-6} \cmidrule(lr){7-7} \cmidrule(lr){8-8}
& GSM8K & ASDiv & SVAMP & College Math & TabMWP & Zebra-grid & CRUX & \\
\midrule
Qwen3-14B-Base & 90.0 & 94.4 & 92.3 & 39.7 & 93.2 & 15.9 & 63.9 & 69.9 \\
Qwen2.5-14B-Ins-1M & \textbf{95.0} & \textbf{95.5} & \underline{94.3} & \underline{42.3} & \underline{96.9} & \underline{20.5} & \textbf{66.4} & \underline{73.0} \\
\midrule
Qwen2.5-7B-Ins-1M & 88.8 & \cellcolor{custompurple}94.7 & 93.0 & 39.6 & 78.4 & 9.2 & 55.6 & 65.6 \\
\model-7B & \cellcolor{custompurple}88.9 & 93.9 & \cellcolor{custompurple}93.6 & \cellcolor{custompurple}39.9 
& \cellcolor{custompurple}89.9 & \cellcolor{custompurple}10.5 & \cellcolor{custompurple}59.0 & \cellcolor{custompurple}68.0 \\
\midrule
Qwen3-8B-Base & 85.2 & 86.6 & 83.3 & 37.6 & 84.0 & 11.9 & 54.1 & 63.2 \\
\model-8B & \cellcolor{custompurple}\underline{93.3} & \cellcolor{custompurple}\underline{95.2} 
& \cellcolor{custompurple}\textbf{94.5} & \cellcolor{custompurple}\textbf{47.1} 
& \cellcolor{custompurple}\textbf{97.2} & \cellcolor{custompurple}\textbf{26.0} 
& \cellcolor{custompurple}\underline{65.3} & \cellcolor{custompurple}\textbf{74.1} \\
\bottomrule
\end{tabular}

\vspace{6pt}

\vspace{2pt}
\textbf{(b) Advanced Mathematics}
\vspace{2pt}

\setlength{\tabcolsep}{1.3pt}
\begin{tabular}{lcccc|c}
\toprule
Models & AIME24 \scalebox{0.8}{avg@64} 
& AIME25 \scalebox{0.8}{avg@64} 
& MATH-500 \scalebox{0.8}{avg@8} 
& Gaokao23en \scalebox{0.8}{avg@8} 
& Avg. \\
\midrule
Qwen3-14B-Base & \underline{14.3} & 10.6 & 77.4 & 64.8 & 41.8 \\
Qwen2.5-14B-Ins-1M$^\dagger$ & 13.1 & \underline{11.2} & \underline{79.0} & \underline{66.0} & \underline{42.3} \\
\midrule
Qwen2.5-7B-Ins-1M & \cellcolor{custompurple}7.6 & 4.2 & 69.6 & 58.3 & 34.9 \\
\model-7B & 7.0 & \cellcolor{custompurple}6.5 & \cellcolor{custompurple}70.9 & \cellcolor{custompurple}61.1 & \cellcolor{custompurple}36.4 \\
\midrule
Qwen3-8B-Base & 11.5 & 10.5 & 64.5 & 56.9 & 35.9 \\
\model-8B & \cellcolor{custompurple}\textbf{23.6} & \cellcolor{custompurple}\textbf{19.9} 
& \cellcolor{custompurple}\textbf{85.5} & \cellcolor{custompurple}\textbf{74.7} 
& \cellcolor{custompurple}\textbf{50.9} \\
\bottomrule
\end{tabular}

\vspace{6pt}
\textbf{(c) Graph Reasoning}
\vspace{2pt}

\setlength{\tabcolsep}{4pt}
\begin{tabular}{lccccc|c}
\toprule
\multirow{2}{*}{Models} & \multicolumn{2}{c}{GraphArena} & \multicolumn{3}{c}{GraphInstruct} & \multirow{2}{*}{Avg.} \\
\cmidrule(lr){2-3} \cmidrule(lr){4-6}
& Poly-Time & NP-Hard & Linear & Poly-Time & NP-Complete & \\
\midrule
Qwen3-14B-Base & 31.6 & 14.8 & 54.3 & 15.5 & 71.8 & 32.6 \\
Qwen2.5-14B-Ins-1M$^\dagger$ & \underline{48.1} & 15.8 & \underline{60.8} & 18.3 & \textbf{81.9} & \underline{40.9} \\
\midrule
Qwen2.5-7B-Ins-1M & 37.2 & \cellcolor{custompurple}15.2 & 46.8 & 10.6 & \cellcolor{custompurple}\underline{75.0} & 32.9 \\
\model-7B & \cellcolor{custompurple}47.7 & 10.7 & \cellcolor{custompurple}51.3 & \cellcolor{custompurple}17.8 & 64.8 & \cellcolor{custompurple}36.2 \\
\midrule
Qwen3-8B-Base & 20.8 & 9.5 & 49.4 & 9.7 & 65.0 & 25.2 \\
\model-8B & \cellcolor{custompurple}\textbf{63.2} & \cellcolor{custompurple}\textbf{32.9} 
& \cellcolor{custompurple}\textbf{64.9} & \cellcolor{custompurple}\textbf{46.3} 
& \cellcolor{custompurple}\underline{75.0} & \cellcolor{custompurple}\textbf{54.4} \\
\bottomrule
\end{tabular}

\end{table}

\subsubsection{Effect on Long CoT Capabilities (RQ3)}
We further investigate whether \model truly incentivizes the Long CoT capabilities of LLMs by examining whether our model exhibits the three core characteristics of Long CoT in complex reasoning problems: \textit{Deep Reasoning}, \textit{Extensive Exploration}, and \textit{Feasible Reflection}.

\paratitle{Deep Reasoning.}
To quantify \textit{deep reasoning}, we measure average response length on challenging mathematics datasets. While response length is not a perfect surrogate for reasoning quality, longer chain-of-thought traces in structured mathematical tasks typically correspond to explicit intermediate deductions and verification steps, making it a practical behavioral proxy for reasoning depth. As shown in Table~\ref{tab:length}, both \model-7B and \model-8B consistently generate much longer responses than their respective base models across all datasets. On average, response length increases by approximately 4.4$\times$ for the 7B backbone and 2.5$\times$ for the 8B backbone, confirming a substantial expansion of reasoning depth after training. Importantly, this increase is not uniform across tasks. The largest multipliers appear on competition-level AIME problems, which demand extended multi-step reasoning, whereas MATH-500 and Gaokao23en exhibit more moderate growth. This task-dependent scaling indicates that our model does not blindly inflate its output, but instead adapts reasoning depth to problem difficulty. Such adaptive allocation of reasoning effort is a hallmark of genuine deep reasoning capability.

\begin{table}[htbp]
\centering
\caption{Response length comparison with the base models (tokens). Cells highlighted indicate the longer ones.}
\label{tab:length}
\centering
\scriptsize
\definecolor{custompurple}{RGB}{208,224,239}
\setlength{\tabcolsep}{5.5pt}
\begin{tabular}{lccccc} 
\toprule
\textbf{Models} & \textbf{AIME24} & \textbf{AIME25} & \textbf{MATH-500} & \textbf{Gaokao23en} & \textbf{Avg.} \\
\midrule
Qwen2.5-7B-Ins-1M & 1933 & 1060 & 711 & 722 & 1107 \\
\model-7B & \cellcolor{custompurple}9013 & \cellcolor{custompurple}6267 & \cellcolor{custompurple}2015 & \cellcolor{custompurple}2137 & \cellcolor{custompurple}4858 \\
\midrule
Qwen3-8B-Base & 3923 & 2620 & 1488 & 1317 & 2337 \\
\model-8B & \cellcolor{custompurple}9106 & \cellcolor{custompurple}8724 & \cellcolor{custompurple}2468 & \cellcolor{custompurple}2692 & \cellcolor{custompurple}5748 \\
\bottomrule
\end{tabular}
\label{tab:length_clean}
\end{table}

\begin{table}[htbp]
\centering
\scriptsize
\caption{Pass@k comparison with the base models. Cells highlighted indicate the better ones.}
\definecolor{custompurple}{RGB}{208,224,239}
\setlength{\tabcolsep}{6pt}
\begin{tabular}{lcccc|c}
\toprule
\textbf{Models} & \textbf{AIME24} & \textbf{AIME25} & \textbf{MATH-500} & \textbf{Gaokao23en} & \textbf{Avg.} \\
\midrule
Qwen2.5-7B-Ins-1M & \cellcolor{custompurple}33.3 & 26.7 & 87.0 & 75.1 & 55.5 \\
\model-7B & 33.0 & \cellcolor{custompurple}33.3 & \cellcolor{custompurple}88.0 & \cellcolor{custompurple}76.9 & \cellcolor{custompurple}57.8 \\
\midrule
Qwen3-8B-Base & 50.0 & 40.0 & 89.4 & 82.6 & 65.5 \\
\model-8B & \cellcolor{custompurple}60.0 & \cellcolor{custompurple}56.7 & \cellcolor{custompurple}93.8 & \cellcolor{custompurple}86.5 & \cellcolor{custompurple}74.3 \\
\bottomrule
\end{tabular}
\label{tab:math_performance}
\end{table}

\begin{figure}[htbp]
    \centerline{\includegraphics[width=\columnwidth]{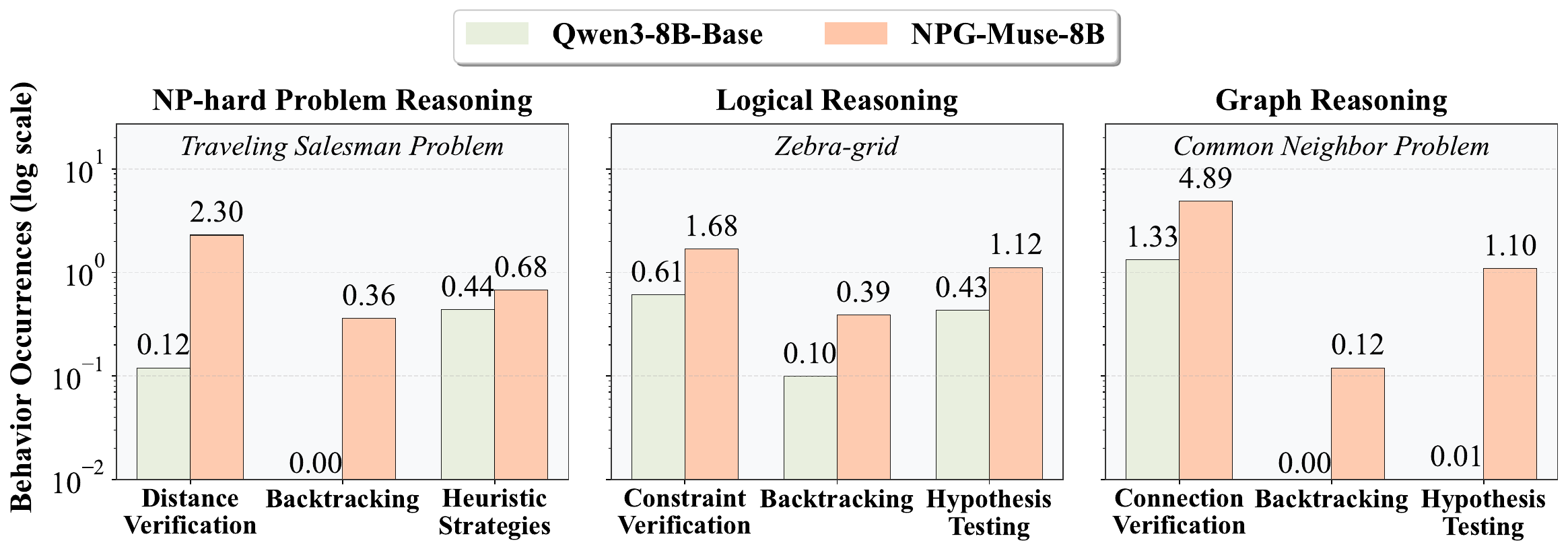}}
    \caption{Self-reflection behavior statistics.}
    \label{fig:cognitive}
    \label{frame}
\end{figure}

\paratitle{Extensive Exploration.} We evaluate \textit{extensive exploration}—the ability to search diverse solution paths—using the \textit{pass@k} metric on challenging mathematics datasets. \textit{pass@k} approximates the model’s upper-bound performance by measuring the probability that at least one of $k$ sampled outputs is correct; consistent gains therefore suggest a broader and more effective exploration of the solution space rather than repeated convergence to the same failure mode. As shown in Table~\ref{tab:math_performance}, both backbones improve after training: the 7B model increases its average pass@k from 55.5 to 57.8, while the 8B backbone rises substantially from 65.5 to 74.3, with particularly large gains on AIME24 (+10.0) and AIME25 (+16.7). Improvements appear across both competition-level and curriculum-style datasets, indicating that enhanced exploration is not task-specific. Together, these results show that the \model-series models expand their effective search over reasoning trajectories, increasing the likelihood of discovering correct solutions under sampling.

\paratitle{Feasible Reflection. } Inspired by \cite{gandhi2025cognitivebehaviorsenableselfimproving}, we adopt an LLM-as-a-judge framework to quantify the frequency of self-reflection behaviors. We select both seen and unseen tasks for observation: TSP (seen during training), Zebra-grid (unseen), and Common Neighbor Problem from Grapharena (unseen). These tasks involve minimal domain-specific knowledge, reducing potential confounding factors and ensuring more reliable statistics. For TSP, we track \textit{verification} (checking correctness), \textit{backtracking} (revising earlier steps), and \textit{heuristics} (using shortcuts or hypothesis testing). For Zebra-grid and Common Neighbor, we retain \textit{verification} and \textit{backtracking}, but replace \textit{heuristics} with \textit{hypothesis testing} to better match the task characteristics. Figure~\ref{fig:cognitive} shows a clear increase in self-reflection across all tasks, demonstrating that our approach substantially enhances the model's ability to refine strategies through self-reflection. Detailed prompts for our LLM-as-a-judge framework are provided in our code, and case studies are in Appendix~\ref{app:transferablity}.

\subsection{Ablation Study}
\begin{table*}[htbp]
\centering
\caption{Ablation on the effect of each training stage. Best accuracy results are highlighted in bold.}
\label{tab:ablation_stage}
\small
\setlength{\tabcolsep}{2.5pt} 
\resizebox{\textwidth}{!}{
\begin{tabular}{l cccccc ccccccccc c}
\toprule
\multirow{2}{*}{\textbf{Models}} & \multicolumn{6}{c}{\textbf{NP-Hard Graph}} & \multicolumn{9}{c}{\textbf{Mathematics}} & \multirow{2}{*}{\textbf{Avg.}} \\
\cmidrule(lr){2-7} \cmidrule(lr){8-16}
& \textbf{TSP(S)} & \textbf{GED(S)} & \textbf{MCP(S)} & \textbf{TSP(L)} & \textbf{GED(L)} & \textbf{MCP(L)} & \textbf{GSM8K} & \textbf{College Math} & \textbf{ASDIV} & \textbf{SVAMP} & \textbf{TabMWP} & \textbf{AIME24} & \textbf{AIME25} & \textbf{MATH-500} & \textbf{Gaokao2023en} & \\ 
\midrule
\multicolumn{17}{l}{\textit{Accuracy (\%):}} \\
Qwen3-8B-Base & 24.8 & 23.4 & 22.4 & 0.0 & 2 & 1.6 & 85.2 & 37.6 & 86.6 & 83.3 & 84.0 & 11.5 & 10.5 & 64.5 & 56.9 & 39.6 \\
+Long-CoT SFT & 76.4 & 58.0 & 84.2 & 0.4 & 3.6 & 23.8 & 82.9 & 43.4 & 81.5 & 80.5 & 84.0 & \textbf{30.6} & \textbf{25.9} & 84.9 & 74.6 & 55.6 \\
+ RL (\model-8B) & \textbf{82.2} & \textbf{63.0} & \textbf{93.6} & \textbf{4.8} & \textbf{65.0} & \textbf{57.2} & \textbf{93.3} & \textbf{47.1} & \textbf{95.2} & \textbf{94.5} & \textbf{97.2} & 23.6 & 19.9 & \textbf{85.5} & \textbf{74.7} & \textbf{62.5} \\
\midrule
\midrule
\multicolumn{17}{l}{\textit{Response length (tokens):}} \\
Qwen3-8B-Base & 2668 & 2864 & 3410 & 3754 & 3126 & 3704 & 664 & 1601 & 607 & 639 & 586 & 3923 & 2620 & 1488 & 1317 & 2198 \\
+Long-CoT SFT & 9925 & 16254 & 5525 & 17418 & 21208 & 12837 & 3633 & 3999 & 4608 & 5307 & 5027 & 12864 & 11874 & 3883 & 4461 & 9255 \\
+RL (\model-8B) & 6218 & 7761 & 4591 & 10282 & 10719 & 8018 & 674 & 1802 & 607 & 909 & 466 & 9106 & 8724 & 2468 & 2692 & 5002 \\
\bottomrule
\end{tabular}
}
\end{table*}

\paratitle{Effect of Each Training Stage.}
We analyze the contribution of each training stage (Long-CoT SFT and RL) to \model-8B's reasoning ability from two perspectives: \textit{effectiveness} (solution accuracy) and \textit{efficiency} (response length). Results are summarized in Table~\ref{tab:ablation_stage}. The Long-CoT SFT stage substantially extends the model's responses (average length increases from $2{,}198$ to $9{,}255$ tokens, a $4.2\times$ increase) and improves performance on most complex reasoning benchmarks, with particularly large gains on NP-hard graph problems and several mathematics datasets. However, SFT also introduces limitations: accuracy decreases on GSM8K, ASDIV and SVAMP, and outputs become excessively long with pronounced repetition on these tasks, indicating reduced robustness and the need for further refinement. After the RL stage, token efficiency is substantially improved while Long-CoT reasoning ability is preserved. The average response length falls to about $5{,}002$ tokens (approximately $2.3\times$ the base model and a 45.9\% reduction relative to the Long-CoT SFT model), and the average accuracy increases from 55.6\% to 62.5\% (a $6.9$ percentage-point gain). In summary, the two-stage training (Long-CoT SFT followed by RL) strikes an effective balance between \textit{reasoning capability} and \textit{token efficiency}.

\paratitle{Effect of Fine-grained Reward Function.}
We analyze the impact of the fine-grained reward by comparing \model-8B with a variant trained using a binary reward during the RL stage, where only fully correct solutions receive a fixed positive reward and all other outputs are left unshaped. As shown in Table~\ref{tab:ablation_reward}, the fine-grained reward increases average accuracy on NP-hard graph problems from 49.2\% to 51.1\%, improves average feasibility from 88.9\% to 91.9\%, and reduces mean response length from 11,910 to 7,931 tokens (a 33\% reduction). As demonstrated in Table~\ref{tab:ablation_reward_2}, these benefits extend to broader reasoning tasks that require strict adherence to complex constraints, including graph, logic, and code reasoning. Overall performance improves consistently across domains, indicating that the fine-grained reward generalizes beyond the training distribution. Collectively, these results suggest that the fine-grained design achieves a better balance between correctness, constraint satisfaction, and conciseness by discouraging repetitive reasoning, explicitly rewarding suboptimal but feasible solutions, and enforcing output-format constraints. This leads to a more reliable ability to solve complex constraint-following problems in practice.

\begin{table}[htbp]
\caption{Ablation on the effect of fine-grained reward on NP-hard graph problems, with better results highlighted in bold.}
\label{tab:ablation_reward}
\small
\resizebox{1\columnwidth}{!}{
\begin{tabular}{l|cccccc|c}
\toprule
\textbf{Model} & \textbf{TSP(S)} & \textbf{GED(S)} & \textbf{MCP(S)} & \textbf{TSP(L)} & \textbf{GED(L)} & \textbf{MCP(L)} & \textbf{Avg.} \\ 
\midrule 
\multicolumn{8}{l}{\textit{Accuracy (\%):}} \\
binary reward & \textbf{83.2} & 57.0 & \textbf{94.6} & 4.2 & 5.6 & 50.6 & 49.2 \\
fine-grained reward & 82.2 & \textbf{63.0} & 93.6 & \textbf{4.8} & \textbf{6.0} & \textbf{57.2} & \textbf{51.1} \\
\hline\hline
\multicolumn{8}{l}{\textit{Feasibility (\%):}} \\
binary reward & 98.8 & \textbf{99.4} & \textbf{98.8} & 71.2 & 81.8 & \textbf{83.4} & 88.9 \\
fine-grained reward & \textbf{99.0} & 99.2 & 97.8 & \textbf{75.2} & \textbf{97.0} & 83.0 & \textbf{91.9} \\
\hline\hline
\multicolumn{8}{l}{\textit{Response Length (tokens):}} \\
binary reward & 11604 & 11335 & 5217 & 17575 & 16184 & 9549 & 11910 \\
fine-grained reward & 6218 & 7761 & 4591 & 10282 & 10719 & 8018 & 7931 \\
\bottomrule
\end{tabular}}
\end{table}

\begin{table}[htbp] 
\centering
\caption{Ablation study of the fine-grained reward design on complex constraint-following tasks.}
\label{tab:ablation_reward_2}
\scriptsize 
\setlength{\tabcolsep}{3.7pt} 
\begin{tabular}{l ccc c c c}
\toprule
\multirow{2}{*}{\textbf{Models}} & \multicolumn{3}{c}{\textbf{GraphInstruct}} & \textbf{Logic} & \textbf{Code} & \multirow{2}{*}{\textbf{Avg.}} \\
\cmidrule(lr){2-4} \cmidrule(lr){5-5} \cmidrule(lr){6-6}
& \textbf{Linear} & \textbf{Poly-Time} & \textbf{NP-Complete} & \textbf{Zebra-grid} & \textbf{CRUX} & \\
\midrule
binary reward& 62.8 & \textbf{47.8} & 62.5 & 25.4 & 59.4 & 51.6 \\
fine-grained reward& \textbf{64.9} & 46.3 & \textbf{75.0} & \textbf{26.0} & \textbf{65.3} & \textbf{55.5} \\
\bottomrule
\end{tabular}
\end{table}
\vspace{-5pt}
\section{Related Work} 

\paratitle{Post-training Techniques for Reasoning Language Models.}
Post-training methods have become essential for enhancing reasoning capabilities in language models. Early techniques like prompt engineering and few-shot learning showed limited effectiveness on complex tasks \cite{10.1145/3560815}. Supervised Fine-Tuning (SFT) offers a more systematic approach by guiding models to imitate reasoning patterns from curated datasets. Combined with rejection sampling \cite{rejectionsampling} for automatic acquisition of high-quality reasoning traces, SFT achieves substantial improvements on challenging benchmarks.

Subsequent methods moved beyond imitation learning. Direct Preference Optimization (DPO) \cite{rafailov2023direct} efficiently aligns models with human preferences but remains constrained by data quality and struggles with complex solution spaces. Reinforcement Learning (RL) proves more powerful, enabling models to explore diverse strategies, receive targeted feedback, and improve beyond fixed datasets. Approaches such as RLHF \cite{ouyang2022training} and PPO \cite{schulman2017proximal} have achieved notable success \cite{lightman2023lets, yuan2023scaling}. Group Relative Policy Optimization (GRPO) \cite{shao2024deepseekmath} further replaces absolute reward estimation with relative comparisons within sampled groups, reducing computational overhead while maintaining stable optimization. This property makes GRPO a practical framework for Long CoT post-training.

\vspace{-2pt}
\paratitle{Language Models for Graph Tasks.}  
Recent progress in graph language models has primarily focused on leveraging LLMs as task-specific solvers. GraphWiz~\cite{chen2024graphwiz} applies instruction tuning with SFT and DPO; GraphThought~\cite{graphthought} introduces thought scaffolding for combinatorial optimization. While effective, these methods rarely investigate how graph reasoning generalizes across domains. More recently, researchers explore graph problems as training corpora for transferable reasoning. NLGift~\cite{nlgift} benchmarks whether graph-trained LLMs acquire generalizable skills; GraphPRM~\cite{graphprm} uses step-wise graph supervision for process reward models; GraphMind~\cite{Zhang2025Improving} employs graph corpora for continued pretraining. Concurrently, G1~\cite{g1} shows reinforcement learning on synthetic graph tasks enhances reasoning without degrading general capabilities. Despite observing transfer beyond graph domains, these works largely hypothesize mechanisms without providing deeper mechanistic evidence of how such transfer occurs.

In contrast, our framework identifies Long CoT capability as a central driver of cross-domain generalization. We demonstrate that NP-hard graph problems naturally elicit the long-horizon reasoning traces essential for transferable logic. By integrating these traces into a our training pipeline, we not only achieve substantial empirical gains but also provide mechanistic evidence for the underlying transfer. This establishes a scalable and robust path toward enhancing the reasoning capabilities of LLMs.

\section{Conclusion and Limitation}

In this work, we introduce NP-hard graph problems as a new training corpus to elicit Long CoT capabilities in LLMs. We propose a two-stage post-training framework combining Long-CoT SFT on rejection-sampled NP-hard graph instances with RL guided by fine-grained reward design. The resulting \model-series models demonstrate significant improvements in Long CoT reasoning, with \model-7B outperforming QwQ-32B on NP-hard graph problems. Our study positions NP-hard graph problems as a scalable and powerful corpus for advancing complex reasoning in LLMs.

Despite these promising results, our work has limitations. The training corpus covers only three types of NP-hard graph problems, and the methodology follows the standard SFT+RL paradigm without exploring alternative post-training strategies. Future work can address these limitations by expanding the problem diversity and investigating broader training approaches.

\newpage
\bibliographystyle{ACM-Reference-Format}
\bibliography{arxiv}

\newpage
\appendix 
\section{More Details of Methodology}
\subsection{Solution Quality Reward}
\label{app:solution}
We apply consistent reward assignments across all tasks: optimal responses receive $2.0$, hallucinated solutions receive $-1.0$, and suboptimal solutions receive task-specific rewards scaled between $0$ and $0.5$ based on the ratio between predicted and optimal values:

\begin{itemize}[leftmargin=*]
    \item \textbf{TSP}: $R_{\textnormal{sub}} = \left(\tfrac{ans}{d}\right)^2 \times 0.5$, where $d$ denotes the predicted tour length.
    \item \textbf{GED}: $R_{\textnormal{sub}} = \tfrac{ans}{c} \times 0.5$, where $c$ denotes the predicted edit cost.
    \item \textbf{MCP}: $R_{\textnormal{sub}} = \tfrac{s}{ans} \times 0.5$, where $s$ denotes the predicted clique size.
\end{itemize}

We apply squaring only for TSP, as its objective values span a broader range than GED or MCP. Near optimality, the raw ratio $\tfrac{ans}{d}$ changes minimally, making fine-grained improvements difficult to capture. Squaring amplifies differences among near-optimal tours while compressing rewards for poor solutions. In contrast, GED and MCP operate within narrower ranges where linear ratios provide sufficient feedback. The $0$--$0.5$ range is chosen to ensure distinguishability from optimal ($2.0$) and hallucinated ($-1.0$) rewards: the lower bound arises naturally from the ratio, while the upper bound of $0.5$—rather than $1.0$—widens the gap between suboptimal and optimal solutions, producing a steeper reward signal that improves the model's sensitivity to solution quality.

\subsection{Repetition Detection Algorithm}
\label{app:repetition}
Our approach (Algorithm \ref{alg:suffix_automaton_repetition}) identifies repetitive patterns by building a suffix automaton from the model's output. The process involves topologically sorting states and calculating occurrence counts through bottom-up propagation to detect meaningful repetitions exceeding predefined thresholds for length and frequency, with prioritization based on frequency, length, and position. With $O(n)$ time complexity, it enables real-time detection even in extensive outputs, making it valuable for reward design in resource-intensive reinforcement learning and for broader applications in production environments and interactive systems.

\begin{algorithm}[htbp]
\small 
\caption{Repetition Detection via Suffix Automaton}
\label{alg:suffix_automaton_repetition}
\begin{algorithmic}[1]
\Require String $S$, minimum length $L_{min}$, minimum repetitions $R_{min}$
\Ensure Detection result $(has\_repetition, substring, count, length, start)$

\State Build suffix automaton $SA$ for $S$ \Comment{$O(|S|)$ time complexity}
\State Topologically sort states by decreasing length
\State Compute cumulative occurrence counts via bottom-up propagation

\State $best \leftarrow null$
\For{each non-initial state $s \in SA$}
    \State $L_{effective} \leftarrow \max(len(s.link) + 1, L_{min})$
    \If{$L_{effective} \leq len(s)$ and $s.occ \geq R_{min}$}
        \State Update $best$ based on prioritization criteria \Comment{Prioritize by frequency, length, and position}
    \EndIf
\EndFor

\State \Return Detection result
\end{algorithmic}
\end{algorithm}

\section{More Experiment Details}
\subsection{Details of Datasets and Metrics}
\label{app:datasets}
Dataset descriptions and evaluation metrics are as follows:

\begin{itemize}[leftmargin=*, topsep=1pt]
\item \textbf{AIME~\cite{aime}. } Competition-level mathematics problems from the American Invitational Mathematics Examination. We evaluate on the 2024 and 2025 problems.

\item \textbf{MATH-500~\cite{math500}. } A curated subset of 500 problems from the MATH benchmark, spanning diverse mathematical domains and difficulty levels.

\item \textbf{GaoKao2023en \cite{gaokao2023}. } An English translation of the mathematics section from the 2023 Chinese National College Entrance Examination (GaoKao). It consists of challenging high-school level problems that require comprehensive logical reasoning. 

\item \textbf{College Math \cite{collegemath}. } A collection of advanced mathematical problems sourced from university-level curricula, including calculus, linear algebra, and probability. Evaluation is based on accuracy.

 \item \textbf{ASDIV~\cite{miao2020asdiv}. } A Diverse Corpus for Evaluating and Developing Math Word Problem Solvers, featuring high linguistic variety and varying mathematical complexity. We evaluate performance using answer accuracy.

\item \textbf{GSM8K~\cite{cobbe2021gsm8k}. } 8.5K grade school math word problems designed to evaluate multi-step numerical reasoning and arithmetic problem-solving. Evaluation uses accuracy.

\item \textbf{SVAMP \cite{patel2021svamp}. } A challenge set of synthetic arithmetic word problems created by applying robust perturbations to existing datasets. It tests the model's reasoning consistency against linguistic variations. Performance is measured by accuracy.

\item \textbf{TabMWP \cite{lu2022tabmwp}. } A large-scale dataset of mathematical word problems where the reasoning context is embedded in tabular data, requiring the model to synthesize information across text and tables. We report the accuracy.

\item \textbf{CRUX~\cite{gu2024cruxevalbenchmarkcodereasoning}.} 800 Python programming problems with input-output test cases, defining input prediction and output prediction tasks. We use the pass@1 metric.

\item \textbf{Zebra-grid~\cite{lin2025zebralogicscalinglimitsllms}.} Logic grid puzzles derived from constraint satisfaction problems, enabling systematic complexity control. Performance is measured by accuracy based on whether solutions satisfy all constraints.

\item \textbf{GraphArena~\cite{tang2024grapharena}.} A benchmark suite with ten graph computational tasks (four polynomial-time, six NP-hard). We measure \textit{accuracy} (optimal solution rate) and \textit{feasibility} (valid solution rate). We utilize the TSP, GED, and MCP subsets for training and in-task evaluation, and the Neighbor and Distance subsets for generalization evaluation.

\item \textbf{GraphInstruct~\cite{chen2024graphwiz}.} An instruction-tuning dataset covering nine graph reasoning tasks with explicit multi-step reasoning paths. We report zero-shot accuracy on the Hamilton Path subset.
\end{itemize}

\subsection{Details of Baselines}
\label{app:baselines}
We incorporate 11 language models as baselines, with detailed descriptions provided below:

\begin{itemize}[leftmargin=*, topsep=3pt, itemsep=1pt]
\item \textbf{Closed-source Models.}
\begin{itemize}[leftmargin=5pt, itemsep=1pt]
\item \textbf{Claude-3.5-Sonnet} \cite{claude35sonnet} is a proprietary model by Anthropic. 

\item \textbf{GPT-4o} \cite{GPT4OpenAI} is a proprietary model by OpenAI. 
\end{itemize}

\item \textbf{Reasoning Models.}
\begin{itemize}[leftmargin=5pt, itemsep=1pt]
\item \textbf{QwQ-32B~\cite{qwq32b}} is a 32B-parameter reasoning model trained with multi-stage RL.

\item \textbf{S1.1-7B} \cite{muennighoff2025s1} is a reasoning model fine-tuned from Qwen2.5-7B-Instruct using SFT based on the s1K-1.1 \cite{muennighoff2025s1} dataset, which includes 1000 questions paired with reasoning traces. 

\end{itemize}

\item \textbf{Non-reasoning Models.}
\begin{itemize}[leftmargin=5pt, itemsep=1pt]
\item \textbf{LLaMA-3-70B} \cite{llama3modelcard} is a 70B-parameter language  model by Meta. 

\item \textbf{Qwen3-8B-Base} \cite{qwen3} is an 8B-parameter base model from Alibaba Cloud, pre-trained on 36 trillion tokens with enhanced reasoning and coding capabilities.

\item \textbf{Qwen2.5-7B-Instruct-1M} \cite{qwen2.5-1m} is a 7B parameter model designed for general-purpose tasks, featuring extended context length up to 1 million tokens.

\end{itemize}

\item \textbf{Graph-oriented Language Models.}
\begin{itemize}[leftmargin=5pt, itemsep=1pt]

\item \textbf{G1-7B} \cite{g1} is a model fine-tuned from Qwen2.5-7B-Instruct, using reinforcement learning on graph-theoretic tasks, outperforming Qwen2.5-72B-Instruct on graph reasoning tasks.

\item \textbf{LLaMA-8B-GT} \cite{graphthought} is a model fine-tuned from LLaMA-3-8B-Instruct to solve graph combinatorial optimization problems. It matches the performance of OpenAI's o1-mini on  GraphArena.

\item \textbf{GraphWiz-7B-DPO} \cite{chen2024graphwiz} is a model fine-tuned from LLaMA-2-7B using SFT and DPO for graph computational problems, outperforming GPT-4 across nine graph tasks.

\end{itemize}
\end{itemize}

\subsection{More Implementation Details}
\label{app:implementation}
\paratitle{Details of Training. }
During training, we utilize synthetically generated small-scale instances of NP-hard graph problems, while evaluation relies on both small- and large-scale instances from GraphArena. To ensure data quality and prevent leakage, we employ GraphArena's official data generation script with different random seeds (43 for SFT and 45 for RL, compared to 0 in the official release) and implement a strict de-duplication pipeline to guarantee the uniqueness of all samples. 

To ensure a balanced distribution across tasks and difficulty levels, we adopt a uniform sampling strategy. During the Supervised Fine-Tuning (SFT) stage, we generate 500 instances per graph size $n \in \{4, \dots, 9\}$ for TSP and GED, and 250 instances per size $n \in \{4, \dots, 15\}$ for MCP, totaling 9,000 training instances. For the Reinforcement Learning (RL) stage, 4-node instances are omitted. We generate 3,000 instances per size for TSP and GED ($n \in \{5, \dots, 9\}$) and 1,500 per size for MCP ($n \in \{5, \dots, 14\}$), yielding a total of 45,000 training instances.
We also design stage-specific prompt templates. In both stages, we modify GraphArena's official templates by removing in-context examples. In the RL stage, we design an extra system prompt which encourages a structured "think first, then answer" approach, discourage code-based solutions, and emphasize step-by-step reasoning. The system prompt template and condensed examples for all three tasks are presented below. 


\begin{exmp}{System prompt in the RL stage}{exmp:system}
\label{appendix:system}
You are a helpful assistant. The assistant first thinks about the reasoning process in the mind and then provides the user with the answer. 
The reasoning process and answer must be formatted exactly as follows:<think>[your reasoning process here]</think><answer>[your answer here]</answer>. 
You can choose to imitate the reasoning process of relevant algorithms or explore an appropriate solution on your own. DO NOT use code, and remember to focus on the logical flow. 
\end{exmp}


\begin{exmp}{Condensed example for TSP}{exmp:tsp}
\label{appendix:tsp_prompt}
You are required to solve the Traveling Salesman Problem for an undirected flight route network. Your objective is to determine the shortest possible route that visits each of the listed airports exactly once and returns to the starting point.\\**Problem to Solve**\\- Airports to visit: BVC, URA, SMR, KOK, MTV, YZF\\- Travel distances (in kilometers) between each pair of airports:\\BVC to KOK: 6401\\URA to KOK: 3138\\SMR to KOK: 11079\\MTV to YZF: 12733\\Please calculate the shortest tour and format your answer as follows: [Airport A, Airport B, Airport C, ..., Airport A]
\end{exmp}

\begin{exmp}{Condensed example for GED}{exmp:ged}
\label{appendix:ged_prompt}
You are required to solve the Graph Edit Distance problem between two molecules. Each edit operation (adding or deleting an edge, adding or deleting an isolated node, or relabeling a node) has the identity cost. Your objective is to establish a mapping between the atom IDs from Molecule A to Molecule B, ensuring that each atom ID in Molecule A corresponds to exactly one atom ID in Molecule B. The mapping corresponds to the minimum edit cost between the two graphs. \\ 
**Problem to Solve**\\
You are given the following two molecules:\\Molecule A:\\- Atoms: Cl (atom 0), C (atom 1), Cl (atom 2), Cl (atom 3).\\- Bonds: 0-1, 1-2, 1-3.\\Molecule B:\\- Atoms: Cl (atom 0), Ge (atom 1), Cl (atom 2), Cl (atom 3).\\- Bonds: 0-1, 1-2, 1-3.\\
Represent the node mapping as a list of integers, where the position in the list corresponds to the atom ID in Molecule A and the value at that position indicates the corresponding atom ID in Molecule B. For instance, if atom 0 in Molecule A corresponds to atom 1 in Molecule B, atom 1 in Molecule A corresponds to atom 0 in Molecule B, and atom 2 remains unchanged, the mapping would be represented as [1, 0, 2, ...].
\end{exmp}



\begin{exmp}{Condensed example for MCP}{exmp:mcp}
\label{appendix:mcp_prompt}
You are required to solve the Maximum Clique Problem for an undirected academic network. In this network, nodes represent authors and edges represent research collaborations. Your objective is to find the largest subset of nodes such that every pair of vertices in this subset is connected by an edge.\\**Problem to Solve**\\- Authors in the network: Gang Zhou, Michel Misson, Pascale Minet, Gary V. Yee, Bhaskar Krishnamachari, Erwan Livolant\\- Research collaborations between these authors: Gang Zhou and Bhaskar Krishnamachari, Michel Misson and Pascale Minet, Michel Misson and Gary V. Yee, Pascale Minet and Bhaskar Krishnamachari, Pascale Minet and Erwan Livolant \\Identify the clique with the maximum number of authors in this network. Present your answer in the following format: [AuthorA, AuthorB, AuthorC, AuthorD, ...].

\end{exmp}

\paratitle{Details of Self-reflection Behavior Evaluation.} 
Inspired by \cite{gandhi2025cognitivebehaviorsenableselfimproving}, we investigate how self-reflection behaviors evolve in the \model models, aiming to uncover the mechanisms underlying their performance improvements on complex reasoning tasks. We focus on three typical types of self-reflection: \textit{verification} (checking correctness), \textit{backtracking} (revising earlier steps), and \textit{heuristics} (using shortcuts or hypothesis testing). To quantify these behaviors, we employ the LLM-as-a-judge framework with Qwen3-8B \cite{qwen3}, a lightweight yet state-of-the-art model. The prompts are adapted to the nature of each task to ensure consistent annotation---for example, heuristics are reframed as hypothesis testing in contexts where this interpretation better reflects the reasoning process. The complete prompts are listed below. 

\begin{exmp}{Prompt template for identifying self-reflection behaviors on TSP task}{exmp:tsp}
\label{appendix:cognitive_tsp_prompt}
1. Distance Verification \\
The following is the reasoning process of a language model solving a TSP (Traveling Salesman Problem) task: [your reasoning process here] \\
Evaluate whether this reasoning process includes steps for verifying distance calculations. Verification steps may include: checking the correctness of distances between two points, verifying total path length calculations, or confirming if a certain path is indeed the shortest. Examples include: "The distance is 116+79+180+189=564" or "Verifying the total distance: 44+114+132+142=432". \\
Count the number of such verification steps and provide this number between the tags <count></count>. If there are no verification steps, return <count>0</count>. \\ 
2. Backtracking Behavior \\
The following is the reasoning process of a language model solving a TSP (Traveling Salesman Problem) task:[your reasoning process here] \\
Evaluate whether this reasoning process includes backtracking behavior, where the model realizes the current path may not be optimal and returns to try different paths. For example, the model might try a path "0->1->2->3->0", find it's not optimal, and then try "0->2->1->3->0". \\
Count the number of distinct backtracking instances and provide this number between the tags <count></count>. If there is no backtracking behavior, return <count>0</count>. \\ 
3. Heuristic Strategies \\
The following is the reasoning process of a language model solving a TSP (Traveling Salesman Problem) task:[your reasoning process here] \\
Evaluate whether this reasoning process uses explicit heuristic strategies, such as the greedy algorithm (always choosing the nearest next city), minimum spanning tree, nearest neighbor algorithm, etc. For example: "Starting from city 0, the nearest city is city 2 with a distance of 44" demonstrates a greedy/nearest neighbor strategy. \\
Count the number of distinct heuristic strategies explicitly used and provide this number between the tags <count></count>. If no heuristic strategies are explicitly used, return <count>0</count>.
\end{exmp}

\begin{exmp}{Prompt template for identifying self-reflection behaviors on common neighbor task}{exmp:neighbor}
\label{appendix:cognitive_neighbor_prompt}
1. Connection Verification \\
The following is the reasoning process of a language model solving a neighbor query task on a graph:[your reasoning process here] \\
Evaluate whether this reasoning process includes steps for verifying connections between nodes. Verification steps may include: checking if edges exist between specific nodes, validating relationships, or confirming the presence of connections. Examples include: "X is connected to Y" or "Verifying that Z is a neighbor of both A and B". \\
Count the number of such verification steps and provide this number between the tags <count></count>. If there are no verification steps, return <count>0</count>. \\
2. Backtracking Behavior \\
The following is the reasoning process of a language model solving a neighbor query task on a graph:[your reasoning process here] \\
Evaluate whether this reasoning process includes backtracking behavior, where the model realizes the current approach may not be correct and returns to try different approaches. For example: "Let me reconsider the connections" or "But let me check if this path is valid". \\
Count the number of distinct backtracking instances and provide this number between the tags <count></count>. If there is no backtracking behavior, return <count>0</count>. \\
3. Hypothesis Testing \\
The following is the reasoning process of a language model solving a neighbor query task on a graph:[your reasoning process here] \\
Evaluate whether this reasoning process includes instances where the model proposes and tests hypotheses about node relationships or graph properties. Examples include: "Let's check if X could be a common neighbor of A and B" or "If we assume this graph is connected, then..." \\
Count the number of distinct hypothesis testing instances where the model makes temporary assumptions and provide this number between the tags <count></count>. If there is no evidence of hypothesis testing, return <count>0</count>.
\end{exmp}

\begin{exmp}{Prompt template for identifying self-reflection behaviors on logic puzzle task}{exmp:logic}
\label{appendix:cognitive_logic_prompt}
1. Constraint Validation \\
The following is the reasoning process of a language model solving a logic puzzle (zebra/Einstein puzzle):[your reasoning process here] \\
Evaluate whether this reasoning process includes steps for validating constraints or checking if a potential solution satisfies the given clues. Examples include: "This placement satisfies clue 3" or "Let's check if this arrangement meets all constraints." \\
Count the number of distinct constraint validation instances and provide this number between the tags <count></count>. If there are no validation steps, return <count>0</count>. \\
2. Backtracking Behavior \\
The following is the reasoning process of a language model solving a logic puzzle (zebra/Einstein puzzle):[your reasoning process here] \\
Evaluate whether this reasoning process includes backtracking behavior, where the model realizes the current path or assumption may not be correct and returns to try different approaches. For example: "This assumption leads to a contradiction, so let's try another possibility" or "Let's reconsider our initial placement for Person X." \\
Count the number of distinct backtracking instances and provide this number between the tags <count></count>. If there is no backtracking behavior, return <count>0</count>. \\
3. Hypothesis Testing \\
The following is the reasoning process of a language model solving a logic puzzle (zebra/Einstein puzzle):[your reasoning process here] \\
Evaluate whether this reasoning process includes instances where the model proposes and tests hypotheses or makes tentative assumptions to explore possibilities. Examples include: "Let's assume Person X is in House Y and see if this works..." or "If we temporarily place the red house at position 3, then..." \\
Count the number of distinct hypothesis testing instances where the model makes temporary assumptions to test outcomes and provide this number between the tags <count></count>. If there is no evidence of hypothesis testing, return <count>0</count>. 
\end{exmp}



\section{More Experiment Results}

\subsection{More Results on Long CoT Capabilities}
\begin{figure}[htbp]
\centerline{\includegraphics[width=\columnwidth]{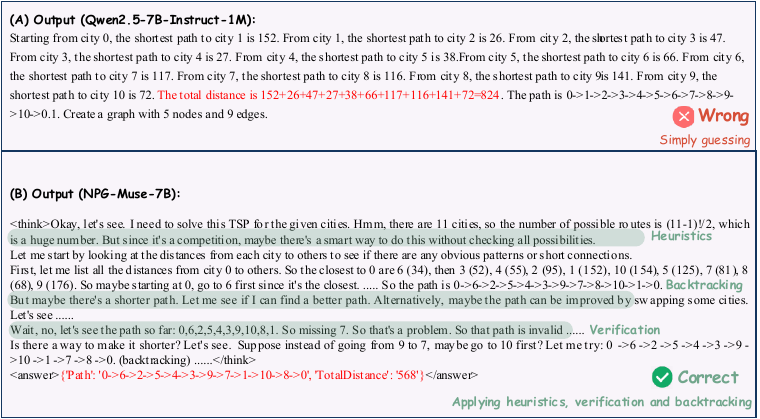}}
\caption{Case study for TSP task.}
\label{fig:tsp-case}
\end{figure}

\begin{figure}[htbp]
\centerline{\includegraphics[width=\columnwidth]{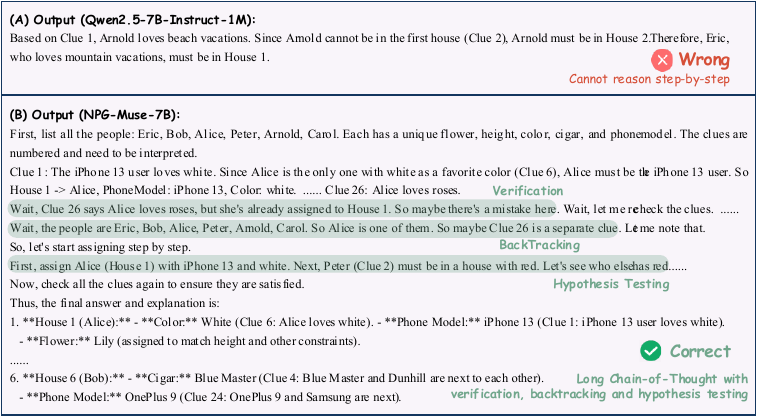}}
\caption{Case study for Logic Puzzle task.}
\label{fig:logic-case}
\end{figure}

As illustrative evidence, Figures~\ref{fig:tsp-case} and~\ref{fig:logic-case} present case studies demonstrating the substantial increase in self-reflection behaviors in \model compared with its base model.

\label{app:transferablity}









\end{document}